\documentclass[runningheads]{llncs}
\usepackage{graphicx}
\usepackage{amsmath}
\usepackage{color}
\usepackage{multirow}
\usepackage{textcomp}
\usepackage{stfloats}
\usepackage{url}
\usepackage{verbatim}
\usepackage{multirow}
\usepackage{graphicx}
\usepackage{algorithm}
\usepackage{amssymb}
\usepackage{wasysym}
\usepackage{booktabs}
\usepackage{colortbl}
\usepackage{subcaption}
\usepackage[misc]{ifsym}
\usepackage{hyperref}

\begin{document}
\title{SimpleFusion: A Simple Fusion Framework for Infrared and Visible Images}
\titlerunning{SimpleFusion}
%
%

%
%


\authorrunning{Ming Chen et al.}
%

\author{Ming Chen\inst{1}$^\dagger$ \and 
Yuxuan Cheng\inst{1}$^\dagger$  \and 
Xinwei He\inst{1}$^{(\textrm{\Letter})}$ \and 
Xinyue Wang\inst{1} \and 
Yan Aze\inst{2}  
\and Jinhai Xiang\inst{1}}
\institute{Huazhong Agricultural University 
\and Huazhong University of Science and Technology \\
\email{\{mchen, hxwxss\}@webmail.hzau.edu.cn, xwhe@mail.hzau.edu.cn} \\
\footnotetext{$\dagger$ Equal contribution. $^{(\textrm{\Letter})}$Corresponding author.}} 
%
\maketitle              
\begin{abstract}

Integrating visible and infrared images into one high-quality image, also known as visible and infrared image fusion, is a challenging yet critical task for many downstream vision tasks. 
Most existing works utilize pretrained deep neural networks or design sophisticated frameworks with strong priors for this task, which may be unsuitable or lack flexibility.
This paper presents SimpleFusion, a simple yet effective framework for visible and infrared image fusion.
Our framework follows the decompose-and-fusion paradigm, where the visible and the infrared images are decomposed into reflectance and illumination components via Retinex theory and followed by the fusion of these corresponding elements.
The whole framework is designed with two plain convolutional neural networks without downsampling, which can perform image decomposition and fusion efficiently.
Moreover, we introduce decomposition loss and a detail-to-semantic loss to preserve the complementary information between the two modalities for fusion.
We conduct extensive experiments on the challenging benchmarks, verifying the superiority of our method over previous state-of-the-arts. Code is available at \href{https://github.com/hxwxss/SimpleFusion-A-Simple-Fusion-Framework-for-Infrared-and-Visible-Images}{https://github.com/hxwxss/SimpleFusion-A-Simple-Fusion-Framework-for-Infrared-and-Visible-Images}
\end{abstract}

\section{Introduction}
Image fusion aims to automatically combine images of distinct but complementary sensors into a high-quality image, which can greatly facilitate extensive downstream applications, such as remote sensing~\cite{simone2002image}, medical imaging~\cite{james2014medical} and video surveillance~\cite{shrinidhi2018ir}. The commonly fused image types include but are not limited to visible, infrared, computed tomography (CT), and magnetic resonance imaging (MRI). Among them, infrared and visible image fusion (IVIF) is a superior research direction due to their ubiquitous sensors (\emph{i.e.}, infrared and RGB sensors) and highly complementary properties. 
Visible images are better at capturing rich appearance information at high spatial resolution, yet they are vulnerable to illumination variation or disguise. Nonetheless, infrared images can naturally complement them by capturing the thermal radiation of the scene. Therefore fusing the two modalies enables in a more robust and accurate perception. 

In general, IVIF can be formulated into a decompose and fusion problem. 
The decomposition step typically decomposes the source images into several components according to signal processing techniques such as multi-scale transform~\cite{liu2015general}, sparse representation~\cite{rubinstein2010dictionaries}, and subspace theory~\cite{van1997subspace}. 
For the second fusion step, it aims to integrate and enhance the corresponding components in the source images to derive a high-quality target one. 


In the past few years, deep learning-based image fusion methods have emerged as a dominant direction in this field.
They typically work by utilizing deep neural networks to decompose features for the source images and then learn to fuse them into high-quality target images. 
Naturally, designing an appropriate framework is essential.
Most works 
utilize pre-trained convolutional neural networks such as VGG19 and ResNet50 for this task.
However, the deep features may dilute the details and may not be a good fit for the low-level fusion task. For low-level tasks, preserving low-level features such as edges, illuminations, and contours is of paramount importance.
Another important research line is to design an auto-encoder architecture for fusion. However, it often involves a handcrafted fusion strategy for better performance.  
Recently, LRRNet~\cite{li2023lrrnet} has developed a sophisticated fusion network guided by low-rank representation. Despite outstanding performance, such an intricate architecture needs to be designed with special care and thus lacks flexibility.

In this work, inspired by Retinex theory, we introduce a simple yet effective framework named SimpleFusion for the infrared and visible image fusion task.
By design, it only consists of two plain two-streamed convolutional neural networks (CNN). 
One two-streamed CNN decomposes the visible image $I$ into reflectance $R$ and illumination $L$ following $I=R \circ L$, where $\circ$ indicates the elementwise product. While the other two-steamed CNN mines corresponding enhancement components from the infrared image to enhance $R$ and $L$ respectively.
The whole framework does not perform feature downsampling and is trained end-to-end, which supports image decomposition and fusion efficiently. 

Our framework has the following merits. First, it intrinsically improves the robustness of image fusion under different lighting scenarios with the Retinex theory. 
Second, it does not perform a downsampling process, thereby fusing the final results to derive the enhanced images is rather natural and flexible, inducing not extra effort for fusion. Moreover, image fusion is a low-level task and keeps the resolution along the convolution layers, reducing low-level detail information loss.  
Lastly, it is simple yet effective. Compared with LRRNet, it simply utilized plain CNN, which is designed with fewer priors on the architecture design. Without bells and whistles, SimpleFusion outperforms existing state-of-the-art methods by a large margin. For instance, on the challenging TNO~\cite{TNO_dataset} dataset, SimpleFusion achieves 6.9045, 89.4448, 13.8089 and 0.10570 on Entropy, Standard Deviation, Mutual Information, and Nabf (the modified fusion artifacts measure), respectively, which are superior to the second-best method LRRNet by a large margin.

To summarize, the contributions of this work are as follows:
\begin{itemize}
    \item We follow Retinex theory and propose to perform visible and infrared image fusions by decomposing the visible images and then learning to mine components in the infrared images to enhance each component, and such a design naturally endows our methods to deal with low-light scenarios.
    \item We present a simple yet effective framework named SimpleFusion, which only adopts plain convolutional neural networks for decomposition and fusion while having fewer priors on the architecture compared with existing works.
    \item  Extensive experiments are conducted on several image fusion benchmarks, demonstrating that SimpleFusion outperforms existing methods by a large margin. 
\end{itemize}

\section{Related Work}

\noindent\textbf{Traditional methods.}
Traditional image fusion methods mainly include weighted average-based fusion~\cite{noushad2017image},transform-domain fusion~\cite{chai2012multifocus}, feature-based fusion\cite{calhoun2008feature} and image pyramid-based fusion~\cite{wang2011multi}. These traditional image fusion methods have their own advantages in various application scenarios and requirements, but they generally suffer from insufficient robustness and are not suitable for complex scenes~\cite{ma2019infrared}.

\noindent\textbf{Deep Learning-based methods.}
With the development of deep learning technology, an increasing number of neural network-based image fusion methods~\cite{li2018densefuse,ma2019fusiongan,zhang2020ifcnn,dong2017cunet,wang2022res2fusion,tang2022ydtr} have begun to receive attention and have achieved excellent results. 
For instance, FusionGAN~\cite{ma2019fusiongan} uses an adversarial framework involving a generator and a discriminator to tackle fusion tasks. Despite impressive fusion results, significant detail loss remains in the outputs. To address this, the authors developed FusionGANv2~\cite{ma2020infrared}, an improved version aimed at enhancing detail preservation. Nonetheless, it encounters challenges with generalization performance. The U2fusion~\cite{xu2020u2fusion} network is designed for multiple fusion tasks. Using the Elastic Weight Consolidation (EWC) algorithm and sequential training strategy, it allows a single model to adapt effectively to various fusion tasks without weight decay. However, the architectural design of the fusion network was not addressed.Architectures based on transformers have also been applied to image fusion tasks. For example, fusion methods like SwinFusion~\cite{ma2022swinfusion} and the YDTR~\cite{tang2022ydtr}. However, the design of these network architectures still requires substantial experimental exploration to discover an excellent fusion network structure.To address this issues, novel approaches have emerged based on the strategy of combining representation models with deep learning, such as CUNet~\cite{dong2017cunet} and LRRNet~\cite{li2023lrrnet}. The network architecture of CUNet~\cite{dong2017cunet} is guided by several optimization problems and multi-modal convolutional sparse coding (MCSC). LRRNet~\cite{li2023lrrnet} is a representation learning guided two-stage fusion network. Its learnable representation model used for source image decomposition exhibits strong interpretability, making image fusion tasks no longer a black art.

\noindent\textbf{Low-light enhancement.}
In 1986, EDWIN H. LAND introduced the retinex theory into the field of image processing, proposing the concept of retinex computation~\cite{land1986alternative}. Until 2004, Zia-ur Rahman and others developed this concept into a comprehensive automated image enhancement algorithm known as Multi-Scale Retinex with Color Restoration (MSRCR)~\cite{rahman2004retinex}. In recent years, the Retinex theory has seen significant development in the field of image enhancement, such as RetinexNet~\cite{retinexnet} and PairLIE~\cite{fu2023learning}. RetinexNet model learns solely through key constraints, including consistent reflectance shared between low-light and normal-light image pairs and smoothness of illumination. Building on this, subsequent brightness enhancement of the illumination is achieved by an enhancement network called Enhance-Net, which also performs joint denoising of reflectance, thus accomplishing image enhancement. PairLIE~\cite{fu2023learning} not only simplifies the network structure and reduces handcrafted priors but also achieves performance comparable to state-of-the-art methods. These low-light image enhancement methods based on the Retinex theory and decomposition ideas have provided us with great inspiration.

\section{Method}
\subsection{Problem Formulation and Challenges}
Given a visible image $I_{v} \in \mathcal{R}^{H\times W \times C}$ and an infrared image $I_{r} \in \mathcal{R}^{H\times W \times 1}$, the objective is to learn a fusion network $f(\cdot)$ which integrates the two sources into a high-quality image $I_{q} \in \mathbb{R}^{H \times W \times C}$ that simultaneously preserves the thermal radiation and rich appearance information, \emph{i.e.}, $I_q = f(I_v, I_r)$. Here $H$, $W$, and $C$ represent the width, height, and the number of channels for the images. 

There are several obstacles to designing an effective fusion framework. (1) The modality gap between visible and infrared images is huge.
Visible images, which are typically composed of three RGB channels, carry rich textural and color information for the scene. However, infrared images have only one-channel robust yet low-contrast thermal radiation about the environment. Therefore, the high incompatibility of the two modalities makes it hard to reconcile them to produce a high-quality output.  
(2) It is difficult to keep the modality-specific information during fusion. Visible and Infrared images have their distinct patterns, these modality-specific properties help describe the same regions of the environment from different perspectives. However, they can be easily lost by disturbance from the other modality during the fusion process. 
(3) Visible images are sensitive to lighting conditions, and it is hard to determine appropriate complementary cues from the infrared modalities for enhancement both efficiently and effectively. 

\subsection{SimpleFusion}
\begin{figure}[t]
\centering
\includegraphics[width=\linewidth]{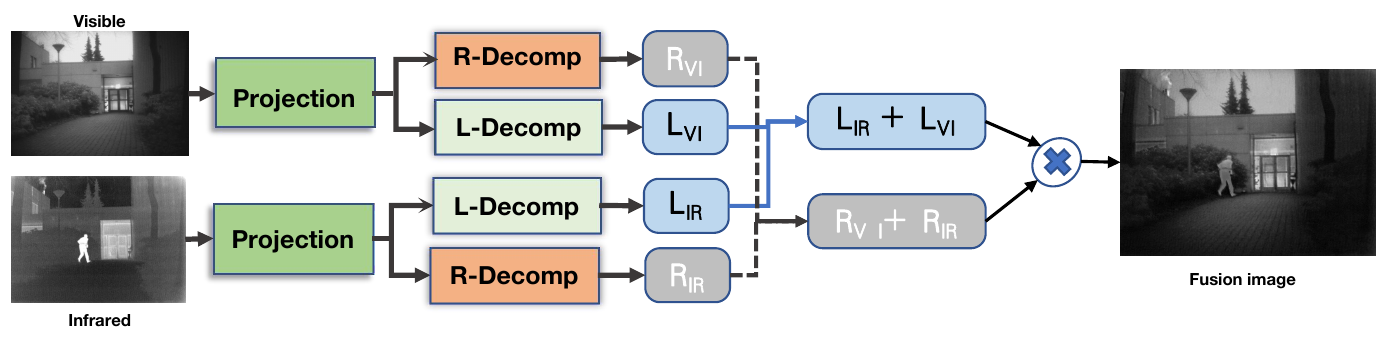}
\caption{The framework of SimpleFusion. It takes infrared and visible images as input, which will be fed into a projection layer to remove unwanted features that are not considered in Retinex theory. For visible image, we decompose it with into illumination and reflectance components, while for infrared image we simply extract corresponding components for fusion, with plain CNNs. The final high-quality image is derived directly by composing the components via Retinex theory. }
\label{framework}
\end{figure}

\noindent\textbf{Overview.} SimpleFusion follows the decompose-and-fusion paradigm. 
As shown in Fig.~\ref{framework}, SimpleFusion is a two-stream framework with one stream decomposing the visible images and the other for infrared images.
Each stream is just the plain convolutional neural network without a resolution reduction layer, thereby the fused image can be naturally derived by directly combining the outputs of the two streams and removing the need to design a specialized decoder.

The decomposing formulation follows the Retinex theory for the visible image, which has been widely adopted in low-light enhancement fields. Given an input visible image $I_v$, it aims to decompose it into illumination component $L$ and reflectance $R$:
\begin{equation}
    I_{v} = L \circ R,
\end{equation}
where $\circ$ represents the element-wise product. 
In our work, we utilize two encoders, denoted by $\Phi_{Ill}(\cdot)$ and $\Phi_{Ref}(\cdot)$ to ensure the decompositions under the following constraints: 
\begin{equation}
    \text{argmin}_{L, R} ||L\circ R - I_{v}|| + \lambda_L \mathcal{L}_{sm}(L) + \lambda_R \mathcal{L}_{sm}(R)
\end{equation}
where $L=\Phi_{Ill}(I_v)$, $R=\Phi_{Ref}(I_v)$ are the estimated illumination and reflectance, and $\mathcal{L}_{sm}$ denotes regularizer which is enforced on the estimated illumination and reflectance, respectively. 

For the corresponding infrared image, we also decompose it into two components, with one enhancing the illumination components while the other enhancing the reflectivity component for the visible image images. The decomposition form is similar to the Retinex decomposition for visible images, except that we treat the visible image as the main modality and the infrared decomposition results are as the supplement.
We simply instantiate another two-stream encoder of the same structure to achieve such a decomposition.

After decomposition, we can simply derive high-quality images by combining the decomposition results. 

\noindent\textbf{Decomposition network.}
Image fusion itself is a low-level task with weak semantic reliance.
Therefore, how to maintain the low-level modality-specific details is essential. 
Previous architecture typically downsamples the images into low-resolution feature maps and then makes a great effort to recover the details by upsampling.
In this paper, we design our decomposition framework by keeping the resolutions along the layers, which greatly facilitates the following fusion process and keep the important local modality-specific information, giving us satisfactory performance.
More specifically, the decomposition network is a two-stream architecture for visible images, where one stream is to estimate illumination components (denoted as Ill-Net), and the other stream (denoted as Ref-Net). 
Each stream is implemented with the same convolutional neural network structure consisting of 5 $3 \times 3$ convolutional layers. 
We utilize ReLU layers as the first four layers. For the last layer, the sigmoid function is leveraged to normalize the outputs to [0, 1]. 
Following Retinex theory, the Ill-Net output is a one-channel illumination map  $L \in \mathbb{R}^{H \times W \times 1}$, and the Ref-Net is a 3-channel output $R \in \mathbb{R}^{H \times W \times 3}$. 
Infrared images contain complementary clues to supplement the visible images to highlight the salient targets.
To achieve this goal, we evaluate the contributions of the infrared images on each component of the visible images.
We utilize an architecture of the same two-stream structure to estimate for enhancement of the illumination and reflectivity, respectively, with one stream producing $\textbf{I}_{i} \in \mathbb{R}^{H \times W \times 3}$ and the other stream producing $\textbf{R}_{i} \in \mathbb{R}^{H \times W \times 1}$. 

\noindent\textbf{Fusion layer.}
After decomposing the visible images and estimating the contributions of infrared images for enhancement, we then fuse them into high-quality images. 
Note that the resolutions are kept during the convolution process, therefore fusing process is rather simple, which is formulated as follows:
\begin{equation}
    \textbf{\textit{I}}_{\text{fusion}} = (\textbf{\textit{L}}_{vi} + \textbf{\textit{L}}_{ir}) \cdot (\textbf{\textit{R}}_{vi} + \textbf{\textit{R}}_{ir})
\end{equation}
SimpleFusion can be seen as a decoder-free network, and eliminate the needs to restore high resolutions from the low-resolution maps, which may dilute the details during the downsampling process. Without the downsampling layers, it can best preserve low-level visual information while also facilitate fusion with minimal effort. 

SimpleFusion is trained to learn to decompose the visible and infrared images and then fuse them into one desired image with improved background details and highlighted targets. 
To this end, it is important to ensure consistency for the decomposition to ensure data fidelity, and at the same time regularize each decomposed component for smoothness. 
We simply follow PairLIE\cite{fu2023learning} and leverage the decomposition loss. Besides, we also follow LRRNet~\cite{li2023lrrnet} and adopt the detail-to-semantic information loss, which can better preserve the complementary information from source images.
These details are elaborated in the following sections. 

\subsection{Decomposition Loss}

Following PairLIE~\cite{fu2023learning},  the decomposition loss includes the Projection term, the Reflectance consistency one, and Retinex one. We describe them below.

\noindent\textbf{Projection loss.}
Retinex decomposition does not consider disturbance components like noise in the image.
Therefore, it is beneficial to remove these useless parts in the image before performing decomposition. 
We simply utilize projection loss, which discards these noise features by projecting the image into another clean one, which is formulated as:
\begin{equation}
    \textbf{\textit{L}}_{{P}} = {{ \|  \textbf{\textit{I}}_ {vi} -  \textbf{\textit{i}}_ {vi} \| }_{2}} \textsuperscript{2}
\end{equation}
where  $\textbf{\textit{i}}_ {{vi}} $ refers to the projected image for input image $\textbf{\textit{I}}_ {vi}$. It helps to transform the raw image into a clean one for decomposition.  

\noindent\textbf{Reflectance consistency loss.}
Reflectance maps that are extracted from the visible images indicate the inherent and invariant physical properties of the objects. We enhance it by incorporating the related components extracted from the infrared images. To this end, it is expected to ensure their matching quality for a better fusion. We further apply consistency loss $\textbf{\textit{L}}_{\text{C}} $ to improve the matching quality, which is formulated as follows:
\begin{equation}
    \textbf{\textit{L}}_{{C}} = {{ \|  \textbf{\textit{R}}_ {vi} -  \textbf{\textit{R}}_ {ir} \| }_{\text{2}}} \textsuperscript{2}
\end{equation}
where $ {R}_{v} $ and $ {R}_{i} $ represent the reflectance maps and the related enhancing components from visible images and infrared images, respectively.

\noindent\textbf{Retinex loss.}
Retinex loss is adopted to ensure the Retinex decomposition. 
Specifically, this loss consists of four terms: the reconstruction loss to ensure data fidelity after reconstruction, two consistency terms for reflectance and illumination, and one smooth term for the initial illumination.
Mathematically, it is defined as follows: 
\begin{equation}
    \textbf{\textit{L}}_{{R}} = {{ \| L \circ R - i \| }_{\text{2}}} \textsuperscript{2} + { \| R -i/stopgrad(L) \|}_{2} \textsuperscript{2} +
    {\| L - {L}_{0} \|}_{2}  \textsuperscript{2} + { \| \triangledown L \|  }_{1}
\end{equation}
where $i$ refers to the projected image, $L_0$ denotes the initial illumination estimation, $\triangledown$ denotes gradients along vertical and horizontal directions. 
According to the above equation, ${{ \| L \circ R - i \| }_{\text{2}}} \textsuperscript{2}$ is the reconstruction term ensuring minor information loss. ${ \| R -i/stopgrad(L) \|}_{2} \textsuperscript{2}$ adds consistency over the estimated reflectances based on the illuminations. Here we detach the gradients from the illuminations for training stability. 
$L_0$ is computed by taking maximum value along the channel dimensions (\emph(i.e.), R,G and B):
\begin{equation}
    \textbf{\textit{L}}_{{0}} = \max_{c \in \{R, G, B\}}  {I^c} (x)
\end{equation}

\noindent\subsubsection {Final Decomposition loss.}
The final decomposition loss function for training our model is given as:
\begin{equation}
    {{L}}_{\text{Decomp}} = \omega_ {{0}} \cdot  \textbf{\textit{L}}_{{P}} + \omega_ {{1}} \cdot  {{L}}_{{C}} + \omega_ {{2}} \cdot  {{L}}_{{R}}
\label{L_Decomp}
\end{equation}
where $ \omega_ {{0}} , \omega_ {{1}},\omega_ {{2}}$ denote the weights. Based on previous works~\cite{fu2023learning}, $ \omega_ {{0}} , \omega_ {{1}},\omega_ {{2}}$ are set to 500, 1, 1 respectively.

\subsection{Detail-to-Semantic Loss}
We follow LRRNet~\cite{li2023lrrnet} and utilize the detail-to-semantic information loss function, which is superior at preserving the complementarity of the visible and infrared images for the fusion process. 
The loss function is computed by exploiting representations from VGG-16~\cite{simonyan2014very} pretrained on ImageNet~\cite{deng2009imagenet}. 

\noindent\textbf{Pixel-level loss.} Compared with the infrared image, the visible image reflects more visual local details. Therefore, we utilize pixel-level loss ${{L}}_{{pixel}}$ to enforce the fused image to have similar visual information as the visible image. Mathematically, it is formulated as follows:
 \begin{equation}
    L_{pixel} = ||I_{\text{fusion}} -  I_{vi}||^{2}_{F}
\end{equation}
where $||\cdot||_{F}$ represents Frobenius norm operation. 

\noindent\textbf{Shallow-level loss.} According to the first convolutional block outputs, we define the shallow-level loss ${L}_{shallow}$, expecting the shallow visual representations of fused images close to that of visible images. The loss is given by:  
\begin{equation}
    {{L}}_{{shallow}} ={\| { \Phi({I}_{fusion}) }\textsuperscript{1} - { \Phi({I}_{vi}) }\textsuperscript{1} \|}_{F} \textsuperscript{2}
\end{equation}
where ${\Phi(\cdot)}\textsuperscript{1} $  represents the first conv-block outputs from the pretarined VGG-16.

\noindent\textbf{Middle-level loss.}
Middle-level loss is calculated based on the features from the second and third convolutional blocks. 
The mid-level features generally reflect perceptual features such as textual and shape information in the images, which are exhibited in both visible and infrared images. Mathematically, it is defined as:
\begin{equation}
    {{L}}_{{middle}} ={\sum_{k=2}^3 \beta ^ k {\| { \Phi({I}_{fusion}) ^ k  - [{w}_{i}  \Phi({I}_{ir}) ^ k  + {w}_{v}  \Phi({I}_{vi}) ^ k ]} \|}_{F} \textsuperscript{2} }
\label{L_middle}
\end{equation}
where $\beta^k$ is the balanced weights for the $k$-the conv-block, $w_{v}$ and $w_{i}$ are the balanced weights for visible and infrared images, respectively.
In practice, $w_{v}$ is set to a smaller value than $w_{i}$ since the visual image is the main modality that contains more visual information. 
We set $w_{v}$ to 0.5 in our framework.

\noindent\textbf{Deep-level loss.} 
We use infrared images to guide the fused images to maintain semantic information. 
Gram Matrix is applied to both infrared and the fused images to extract such information. 
The loss function ${L}_{deep}$ is defined as follows:
\begin{equation}
    {{L}}_{{deep}} = {\| {Gram({\Phi({I}_{fusion}) ^ 4})} - {Gram({\Phi({I}_{ir}) ^ 4})} \|}_{F} \textsuperscript{2}
\end{equation}

The final detail-to-semantic loss is constructed as follows: 
\begin{equation}
    {{L}}_{\text{D2S}} = {\gamma_{1}} \cdot  {{L}}_{{pixel}} +  {\gamma_{2}} \cdot  {{L}}_{{shallow}} +  {{L}}_{{middle}} + 
     {\gamma_{4}} \cdot  {{L}}_{{deep}}
\label{L_d2s}
\end{equation}
where ${\gamma_{1}}, {\gamma_{2}}, {\gamma_{4}}$ are the balanced weights. 
Note that for the low-level image fusion task, the local details are more important and should be set to a larger weight. Therefore, we set $ {\gamma_{1}}$ to 10 to preserve more local details. 

\subsection{Overall Loss Function}

We combine the decomposition and the detail to semantic losses to train our framework:
\begin{equation}
    {{L}}_{{total}} = \lambda * {L}_{\text{Decomp}} + {L}_{\text{D2S}}
\end{equation}
where $\lambda$ balances the magnitude difference between the decomposion and detail to semantic loss functions. We empirically set it to 1000 for better results.  

\section{Experiments}
\subsection{Experimental Setups}
\noindent\textbf{Datasets.}
Following previous works~\cite{li2023lrrnet}, our approach leverages the KAIST~\cite{hwang2015multispectral} dataset, which comprises 95,328 pairs of infrared-visible light images. We randomly selected 20,000 pairs from this dataset as our training set. Additionally, we have combined two public datasets to create a robust test set. Specifically, the test set is composed of 21 pairs of data from the TNO~\cite{TNO_dataset} test set and an additional 40 pairs of data from the VOT2020-RGBT~\cite{kristan2020eighth} dataset. This combination provides a diverse and extensive set of data for evaluating the performance of our framework.

\noindent\textbf{Implementation details.}
We implement SimpleFusion with PyTorch and perform optimization with ADAM. The learning rate is set to $1 \times 10^{-5}$.
We randomly select 20,000 pairs of images from the KAIST~\cite{hwang2015multispectral} dataset as training data, with input images converted to gray and compressed to $128 \times 128$. The model is trained on a single NVIDIA RTX 3090, using a batch size of 8 for 4 epochs.

\noindent\textbf{Evaluation metrics.}
To evaluate our model, a comprehensive set of four quantitative metrics has been employed, which encompasses Entropy (En), Standard Deviation (SD), Mutual Information (MI), and the modified fusion artifacts measure (Nabf). 
For these metrics, the higher the values, the better (except Nabf).

\subsection{Comparisons with State-of-the-arts}

We compare our method with 10 representative image fusion frameworks: an encoder-decoder based method DenseFuse~\cite{li2018densefuse}, a GAN based method FusionGAN~\cite{ma2019fusiongan}, a CNN-based general framework IFCNN~\cite{zhang2020ifcnn}, an ISTA-based algorithm CUNet~\cite{dong2017cunet}, a residual fusion network RFN-Nest~\cite{li2021rfn}, a Res2Net-based algorithm Res2Fusion~\cite{wang2022res2fusion}, a transformer-based framework YDTR~\cite{tang2022ydtr}, a Swin-transformer-based method SwinFusion~\cite{ma2022swinfusion}, a unified fusion network U2Fusion~\cite{xu2020u2fusion}, and a representation learning guided fusion network LRRNet~\cite{li2023lrrnet}.

\noindent\textbf{Fusion results on TNO.}
Table~\ref{tab:metrics_comparison1} summarizes the comparison results with existing state-of-the-art methods on TNO. As shown, SimpleFusion achieves the best scores across three metrics (EN, SD and MI), particularly with a significant improvement in SD. In terms of Nabf, we obtain competitive performance when compared with existing state-of-the-arts, suggesting that the image exhibits a large spatial variation in grayscale values, resulting in higher pixel contrast and richness of detail and contrast.

\begin{table*}[t]
\centering
\setlength{\tabcolsep}{2.6mm}
\begin{tabular}{l|c|ccccc}
\hline
Method &Year & En (↑) & SD (↑) & MI (↑)   & Nabf (↓) \\
\hline
DenseFuse~\cite{li2018densefuse} & 2019 & 6.67158 & 67.57282 & 13.34317  & 0.09214 \\

FusionGAN~\cite{ma2019fusiongan}  & 2019 & 6.36285 & 54.35752 & 12.72570   &  \textcolor{blue}{0.06706} \\

IFCNN~\cite{zhang2020ifcnn} & 2020 & 6.59545 & 66.87578 & 13.19090 &  0.17959 \\

CUNet~\cite{dong2017cunet} & 2020 & 6.13996 & 43.53543 & 12.27992 & 0.16574 \\

RFN-Nest~\cite{li2021rfn} & 2021 & 6.84134 & 71.90131 & 13.68269  &  0.07288 \\

Res2Fusion~\cite{wang2022res2fusion} & 2022 & 6.67774 & 67.27749 & 13.35549  &  0.09223 \\

YDTR~\cite{tang2022ydtr} & 2022 & 6.22681 & 51.48819 & 12.45363  & \textcolor{red}{ 0.02167} \\

SwinFusion~\cite{ma2022swinfusion} & 2022 & 6.68096 & 80.41930 & 13.36191  & 0.12478 \\

U2Fusion~\cite{xu2020u2fusion} & 2022 & 6.75708 & 64.91158 & 13.51416 &  0.29088 \\

LRRNet~\cite{li2023lrrnet} & 2023 & \textcolor{blue}{6.85836} &  \textcolor{blue}{81.78905} &  \textcolor{blue}{13.71673}  & 0.14168 \\
\hline
SimpleFusion & - &\textcolor{red}{6.90455} &\textcolor{red}{89.44478} & \textcolor{red}{13.80891}  & 0.10570 \\ 
\hline

\end{tabular}
\caption{Fusion results on TNO. \textcolor{red}{Red} and \textcolor{blue}{Blue} indicate the best and the second best, respectively}
\label{tab:metrics_comparison1}
\end{table*}

\begin{table*}[t]
\centering
\setlength{\tabcolsep}{2.6mm}
\begin{tabular}{l|c|ccccc}

\hline
Method &Year & En (↑) & SD (↑) & MI (↑)  & Nabf (↓) \\
\hline
DenseFuse~\cite{li2018densefuse}  & 2019 & 6.77630 & 73.63462 & 13.55261 & 0.06346 \\
FusionGAN~\cite{ma2019fusiongan} & 2019 & 6.52031 & 62.84940 & 13.04062  & 0.07527 \\
IFCNN~\cite{zhang2020ifcnn}  & 2020 & 6.74105 & 76.24922 & 13.48210 &  0.20119 \\
CUNet~\cite{dong2017cunet}  & 2020 & 6.33359 & 49.71923 & 12.66718 &  0.19043 \\
RFN-Nest~\cite{li2021rfn}  & 2021 & 6.92952 & 78.22247 & 13.85904 & 0.06357 \\
Res2Fusion~\cite{wang2022res2fusion}  & 2022 & 6.78124 & 73.61685 & 13.56248  & \textcolor{red}{0.06372} \\
YDTR~\cite{tang2022ydtr}  & 2022 & 6.40119 & 62.44826 & 12.80238   & 0.02648 \\
SwinFusion~\cite{ma2022swinfusion}  & 2022 & 6.81625 & 89.41668 & 13.63250   & 0.14224 \\
U2Fusion~\cite{xu2020u2fusion}  & 2022 & 6.94865 & 76.78378 & 13.89730  & 0.28297 \\
LRRNet~\cite{li2023lrrnet} & 2023 & \textcolor{red}{6.97205} & {89.05225} & \textcolor{red}{13.94410} & 0.13162 \\
\hline
SimpleFusion & - &6.70115 &\textcolor{red}{95.64936} &13.40232 &0.09867\\

\bottomrule
\end{tabular}
\caption{Fusion results on VOTRGBT-TNO.\textcolor{red}{Red} 
indicates the best}
\label{tab:VOTRGBT-TNO}
\end{table*}
\noindent\textbf{Fusion results on VOTRGBT-TNO.} 
Following LRRNet~\cite{li2023lrrnet}, 40 pairs of images are selected from
VOT2020-RGBT~\cite{kristan2020eighth} and TNO~\cite{TNO_dataset} to construct a new test dataset. 
According to quantitative results in Table~\ref{tab:VOTRGBT-TNO}, we can observe that SimpleFusion further improves the SD metric on this diverse dataset, significantly outperforming previous methods. Note that a higher SD (standard deviation) in an image indicates that the variation or distribution of pixel values within the image is more extensive or diverse.
These performance improvements on this metric manifest richer and more diverse details for the fused images, which may facilitate downstream feature extraction and further analysis.
\begin{table}[!t]
\centering
\setlength{\tabcolsep}{4.6mm}
\begin{tabular}{l|c|ccccc}
\hline
$\gamma_{2}$ &$w_{i}$ & En (↑) & SD (↑) & MI (↑)  & Nabf (↓) \\
\hline
\multirow{4}{*}{0.1} & 2.0  & 6.8259& 86.9452  & 13.6519  & \textcolor{red} {0.09936}\\
& 3.0  & 6.7305& 75.9124  & 13.4610 & 0.11568\\
& 4.0 & 6.5218 & 56.2967 & 13.0437  & 0.12035 \\
& 5.0  & 6.5681 & 61.0515 & 13.1363 & 0.19352\\
\hline 
\multirow{4}{*}{0.5 } & 2.0  & 6.8745 &87.1859  & 13.7490  &0.10106 \\
& 3.0  & 6.7480&75.5038 & 13.4959 & 0.10941\\ 
& 4.0 & 6.6813& 67.3020 & 13.3627  & 0.11983 \\
& 5.0  & 6.4801 & 58.2438  & 12.9603 & 0.17948\\
\hline
\multirow{4}{*}{1.0} & 2.0  & 6.8961 & 89.4250  & 13.7921 & 0.10431 \\ 
& 3.0  & 6.7857 &  80.6186  & 13.5715 & 0.09281\\ 
& 4.0 &  6.7052 & 70.8964 & 13.4104  &  0.11244 \\
& 5.0  & 6.6590 & 62.8026 & 13.3180  &  0.12652\\  
\hline
\multirow{4}{*}{1.5} & 2.0  &\textcolor{red} {6.9093} & \textcolor{red} {90.6219}  & \textcolor{red} {13.8187} & 0.15237 \\  
& 3.0  & 6.8616 &  84.3247  & 13.7232& 0.12095\\  
& 4.0 &  6.7690 & 74.4652 &  13.5380  &  0.11099 \\   
& 5.0  & 6.7401 & 71.9136 & 13.4802  &   0.12004\\  
\hline
\multirow{4}{*}{2.0} & 2.0  & 6.8988  &  87.8578  & 13.7977 & 0.11372 \\
& 3.0 & 6.8562 & 84.4729 & 13.7124 & 0.10162\\
& 4.0  & 6.7863  & 77.8587 &  13.5726  &  0.10537\\ 
& 5.0 & 6.7459 & 70.5723 & 13.4918 & 0.12795\\ 
\hline
\multirow{4}{*}{2.5} & 2.0  &\textcolor{blue}{6.9045} &\textcolor{blue}{89.4448} & \textcolor{blue}{13.8089}  & \textcolor{blue} {0.10570} \\
& 3.0 & 6.8797  & 84.1515 & 13.7595 & 0.12733 \\  
& 4.0  &  6.8248 &  80.8176 & 13.6496 & 0.10997\\ 
& 5.0  &  6.7738 & 75.9598 & 13.5477 & 0.10895\\ 
\hline
\end{tabular}
\caption{Impact of $\gamma_{2}$ and $w_{i}$ on TNO. \textcolor{red}{Red} indicates the best, and \textcolor{blue}{Red} indicates the most balanced parameter combination.}
\label{tab:Ablation study}
\end{table}

\subsection{Ablation study}
\begin{figure}[!h]
\centering
\includegraphics[width=.8\linewidth]{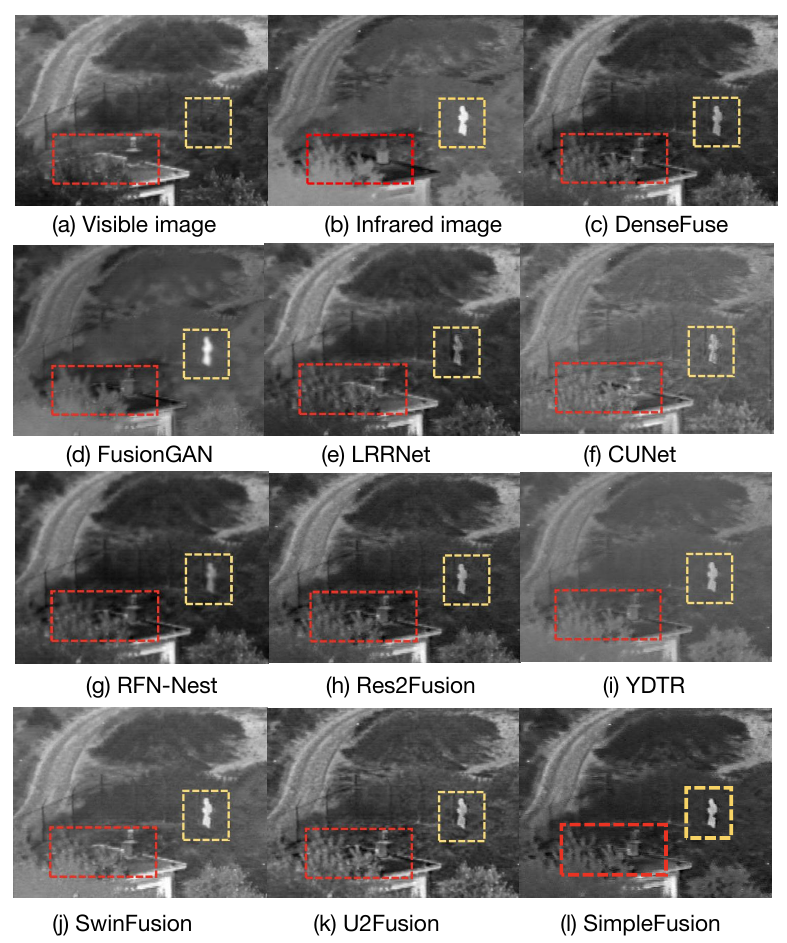}
\caption{The typical fusion results on TNO (``man'' image).}
\label{visiablization_1}
\end{figure}
\noindent\textbf{Impact of $\gamma_{2}$ and $w_{i}$.} The loss functions involve a set of hyperparameters to be tuned. 
In this section, we mainly investigate the impact of hyper-parameters $\gamma_{2}, \gamma_{4}$ and $w_{i}$. While for ($\omega_0, \omega_1, \omega_2,\gamma_{1}$ and $w_{v}$) in Eq.~\ref{L_Decomp}, Eq.~\ref{L_middle} and Eq.~\ref{L_d2s}, we empirically set them according to~\cite{fu2023learning,li2023lrrnet}. 
Our ablation experiments are summarized in Tab.~\ref{tab:Ablation study}.
As shown, when $\gamma_{2} = 1.5 $ and $ {w}_{i} = 2.0 $, our model obtains the best in terms of En, SD and MI. However, at the same time, our model performs the worst on the metric Nabf. 
It implies that the fused image contains excessive noise and is visually perceived as unnatural. 
When $\gamma_{2} = 0.1 $ and $ {w}_{i} = 2.0$, our model reaches the best scores on Nabf. However, it performs poorly on the other metrics.
Overall, SimpleFusion has a satisfactory performance across all metrics when $\gamma_{2} = 2.5 $ and $ {w}_{i} = 2.0 $, which are our default configurations in all our following experiments. 

\noindent\textbf{Visualization.} 
Fig.~\ref{visiablization_1} compares typical fusion results of different methods. Observing the red box in Fig.\ref{visiablization_1}, subjective evaluations show that fusion images generated by methods such as CUNet, YDTR, and SwinFusion appear blurry and lack texture details. On the other hand, methods like DenseFuse, FusionGAN, Res2Fusion, U2Fusion, and LRRNet preserve textures but may introduce noise into the fused images. Observing the yellow box in Fig.\ref{visiablization_1}, subjective evaluations show that fusion images generated by methods such as DenseFuse,  CUnet, RFN-Nest, Res2Fusion, TDTR, U2Fusion, and LRRNet appear to significantly lack the features of the target(the “man”).In contrast, in the fused images generated by IFCNN, FusionGAN, and SwinFusion, the features of the target are very prominent, but the edge transitions still lack sharpness. In the images produced by our SimpleFusion method, the target features are prominent and the transitions at the image edges are sharp enough. Furthermore, the output from our fusion network yields a more natural-looking image.

\section{Conclusion}

In this paper, we have presented a simple yet effective image fusion framework for visible and infrared images. 
Compared with existing works, our framework only adopts plain convolutional neural networks with much fewer priors in the architecture design, thereby being more flexible. 
In our framework, for the visible images, a two-stream CNN is utilized to decompose it into illuminance and reflectance. For infrared images, we calculate the related components to enhance illuminance and reflectance, respectively. Our whole framework keeps the resolution along the layers which supports fusing each component with minor efforts. 
Extensive experiments have been done to prove its superiority. 
However, our framework has many hyperparameters in the loss for tunning. In the future, we plan to adaptively just them in our framework instead of manually tuning them.
\section{Acknowledgement}
This work is supported by the National Natural Science Foundation of China (No.62302188); Hubei Province Natural Science Foundation (No.2023AFB267); Fundamental Research Funds for the Central Universities (No.2662023XXQD001).
%
%
%
\bibliographystyle{splncs04}
\bibliography{paper}

\end{document}